\documentclass[utf8]{mod_FrontiersinHarvard} 

\pdfoutput=1
\usepackage{url,hyperref,lineno,microtype,subcaption}
\usepackage[onehalfspacing]{setspace}
\graphicspath{{figures/}}

\usepackage{algpseudocode}
\usepackage{algorithm}

\def\keyFont{\fontsize{8}{11}\helveticabold }
\def\firstAuthorLast{Orchard {et~al.}} 
\def\Authors{Jeff Orchard\,$^{1,*}$, Russell Jarvis\,$^{2}$}
\def\Address{$^{1}$Cheriton School of Computer Science, University of Waterloo, Waterloo, ON, Canada \\
$^{2}$International Centre for Neuromorphic Systems, Western Sydney University, Penrith, NSW, Australia }
\def\corrAuthor{Jeff Orchard}

\def\corrEmail{jorchard@uwaterloo.ca}

\def\v{v}
\def\V{\mathbb{V}}
\def\bind{\otimes}
\def\unbind{\oslash}
\def\sim{\odot}
\def\bundle{\oplus}
\def\permute{\rho}
\def\thresh{\theta}
\def\Complex{\mathbb{C}}
\def\Real{\mathbb{R}}
\def\e{\mathrm{e}}
\def\i{\mathrm{i}}
\newcommand{\mye}[1]{\e^{\i #1}}
\newcommand{\myen}[1]{\e^{-\i #1}}

\begin{document}
\onecolumn
\firstpage{1}

\title[HD Computing with Phasors]{Hyperdimensional Computing with Spiking-Phasor Neurons} 

\author[\firstAuthorLast ]{\Authors} 
\address{} 
\correspondance{} 

\extraAuth{}

\maketitle

\begin{abstract}
Vector Symbolic Architectures (VSAs) are a powerful framework for representing compositional reasoning. They lend themselves to neural-network implementations, allowing us to create neural networks that can perform cognitive functions, like spatial reasoning, arithmetic, symbol binding, and logic. But the vectors involved can be quite large, hence the alternative label \emph{Hyperdimensional (HD) computing}. Advances in neuromorphic hardware hold the promise of reducing the running time and energy footprint of neural networks by orders of magnitude. In this paper, we extend some pioneering work to run VSA algorithms on a substrate of spiking neurons that could be run efficiently on neuromorphic hardware.
\section{}

\tiny
 \keyFont{ \section{Keywords:} VSA, hyperdimensional computing, phasor, spiking, neuromorphic, resonate-and-fire} 
\end{abstract}

\section{Introduction}

Vector Symbolic Architectures (VSAs) offer a way to represent symbols (such as objects or properties) using high-dimensional vectors. Also called Hyperdimensional (HD) computing, these vectors can be combined in different ways to produce other vectors in the same vector space. Amazingly, those new vectors can still be decomposed into their constituent vectors. Hence, a VSA forms a type of compositional language over a set of symbols (called a \emph{vocabulary}), and is proposed as the key to modelling cognitive processes in the brain \citep{Gayler2003}. For a survey of VSA approaches, see \citep{Schlegel2022}.

VSAs offer a facility to model cognitive processes within a neural-network substrate. Some implementations use artificial neurons \citep{Pale2022}, and some use spiking-neuron models \citep{Stewart2011a,Frady2018}. However, the dimensionality of these vectors can be quite large.

Neuromorphic hardware offers a computing platform that can run neural networks at a fraction of the energy cost and running time, compared to other hardware \citep{Thakur2018}. The prospect of running VSA algorithms on neuromorphic hardware seems like an ideal pairing.

To take full advantage of neuromorphic hardware, a spiking VSA needs to be developed. In this paper, we extend the work of \cite{Frady2019}, and implement a full VSA suite in spiking neurons.

\subsection{Vector Symbolic Architectures}


A Vector Symbolic Architecture (VSA) is a vector space paired with a small collection of operations: similarity $\sim$, binding $\bind$, unbinding $\unbind$, bundling $\bundle$, permutation $\permute(\cdot)$, and clean-up.

The {\bf similarity} operation measures how ``close'' two vectors are. In a VSA, randomly-chosen, high-dimensional vectors tend to yield a similarity close to zero. For example, consider a vector space $\V$, and two random vectors, $u,v \in \V$. We would expect that $u\sim v \approx 0$, while $v\sim v = u\sim u = 1$.

{\bf Binding} combines two vectors to get another vector that is not similar to either of the two. Thus, if $w = u\bind v$, then $w\sim u \approx 0$ and $w \sim v \approx 0$. Unbinding does the opposite of binding, so that $w \unbind u \approx v$.

{\bf Bundling} creates a vector that is still somewhat similar to the two constituent vectors. If $w=u\bundle v$, then $w\sim u \gg 0$, though it will likely not be close to 1. This corruption comes from the fact that some information is lost in the bundling operation. How much is lost depends on the vector space, the specific bundling operation, and how many vectors were bundled together.

{\bf Permutation} jumbles a vector -- in a reversible way -- so that the resulting vector is dissimilar from the original. Thus, if $w = \permute (v)$, then $w \sim v \approx 0$, but $\permute^{-1} (w)$ yields $v$ back.

Finally, {\bf clean-up} takes a vector and restores it to the closest match in the VSA's vocabulary. The clean-up is helpful because the operations of binding, bundling, etc., tend to corrupt the vectors, and noise can accumulate. Using the noisy vectors in further operations could start to affect their proper function and produce unpredictable behaviours. The clean-up operation can undo that corruption and the clean vector can be used with greater confidence.

\subsection{Fourier Holographic Reduced Representation}

Fourier Holographic Reduced Representation (FHRR) is a VSA that uses complex-valued unitary vectors to encode symbols. Let $\v \in \Complex^N$ be a complex vector. The vector $\v$ is \emph{unitary} if $| \v_k | = 1$ for $k=1, \ldots , N$. Using exponential notation, we can write $\v_k$ as $\mye{\phi_k}$, where $\phi_k$ is the phase of the complex number. FHRRs are related to Holographic Reduced Representations (HRRs) \citep{Plate1995}. If the vectors in the FHRR exhibit conjugate symmetry (so that $\phi_k = -\phi_{N-k}$), then taking the inverse Fourier transform of $v$ yields a real-valued vector, $\bar{v} \in \Real^N$.

{\bf Binding:} In an FHRR, the binding operation is done using element-wise multiplication, the \emph{Hadamard} product. Since the elements of the FHRR vectors are unit-modulus complex numbers, multiplying them is equivalent to adding their phases,
$$
\mye{\phi_1} \, \mye{\phi_2} \ = \ \mye{\left( \phi_1 + \phi_2 \right) } \ .
$$
The inverse operation of unbinding is elementwise division (or multiplication by the conjugate), as in
$$
\mye{\phi_1} \, \myen{\phi_2} \ = \ \mye{\left( \phi_1 - \phi_2 \right) }
$$

{\bf Similarity:} If $u$ and $v$ are vectors in an FHRR, then their similarity, $u \sim v$, is computed using the complex inner product, $u \cdot {\bar v}$, where $\bar{v}$ is the conjugate of of $v$.

{\bf Bundling:} In most VSAs, including FHRR, the bundling operation is simply vector addition. Consider adding two unit-modulus complex numbers, $\mye{\phi_a} + \mye{\phi_b}$. The resulting complex number will have a phase of $\frac12 (\phi_a+\phi_b)$, but will probably not be unit-modulus. In the FHRR, the modulus is ignored, and only the phase is kept. Discarding the modulus is a source of information loss, and is one of the reasons that bundling has limitations.

{\bf Permutation:} Permutation is done by literally permuting the vector elements. If ${\bf P}$ is a permutation matrix, then $\rho (v) = {\bf P} v$, and $\rho^{-1} (u) = {\bf P}^{-1} (u)$.

{\bf Clean-up:} In an FHRR, clean-up is often done either by an attractor network (in which the vocabulary vectors are fixed points) \citep{Frady2018,Frady2019,Frady2022}, or by an associative memory \citep{Stewart2011}.

{\bf Fractional Binding:} The FHRR has an additional operation that is not present in all VSAs. Consider binding a vector $v$ with itself, yielding $v \otimes v$. In the FHRR, $\bind$ is the Hadamard product, so you can write that as $v^2$, where the exponent is applied elementwise. But now one can contemplate using non-integer exponents, such as $v^{1.5}$, or $v^{0.7}$. This is called \emph{fractional binding}, or \emph{Fractional Power Encoding} (FPE) \citep{Plate1995,Frady2022}. If one of the elements of our unitary vector is $\mye{\phi}$, with $-\pi < \phi \leq \pi$, then raising that vector to the exponent $\alpha$ yields a vector with the element $\mye{\phi \alpha}$. In other words, the fractional power encoding of $\alpha$ is the same as multiplying the phase by $\alpha$.

\section{Methods}

As pointed out in \citep{Frady2019}, the phase of a phasor can be represented by the timing of a spike within a cycle. For example, the phase of the complex number $\mye{\frac{\pi}{2}}$ can be represented as a spike occurring 0.25 seconds into a cycle that has a period of 1 second, since the ratio $0.25:1 = \frac{\pi}{2}:2\pi$. We will call these neurons \emph{spiking phasors}. In a periodic spiking system, the phase angle, $\phi \in [0, 2\pi]$, and the spike offset, $t\in [0,T]$, are directly related by the spiking frequency, $\lambda$ Hz, according to
$$
\frac{t}{T} = \frac{\phi}{2\pi} \ , \quad \text{where} \ \ T = \frac1\lambda \ .
$$
Thus, $\phi = \frac{2\pi t}{T}$ converts the spike time, $t$, to phase angle $\phi$. Throughout this paper, we will tend to use phase angles when we refer to spike times, but with the understanding that they are isomorphic to each other.



In the following sections, we describe a series of neuron models that perform the spike-timing computations that correspond to the basic FHRR vector operations.

\subsection{Phase-Sum Neuron Model}

Recall that binding two unitary vectors involves taking a Hadamard product of the form $\mye{\phi_a} \mye{\phi_b}$, which is $\mye{(\phi_a + \phi_b)}$. Thus, binding in the FHRR is the same as phase summation. In this section, we describe a model for a spiking neuron such that the phase of its spikes occur at the sum of the phases of the incoming spikes. 

Firstly, this neuron (and most other neuron models we will describe) has an integrator variable, $x$, that simply keeps track of the global phase. Recall that these spiking phasor networks are synchronized to a global cycle. If we make $x$ follow the differential equation
$$
x' = 1 \ , \quad x(0) = 0 \ ,
$$
then $x(2\pi) = 2\pi$, signifying the end of the cycle, at which point $x$ is reset back to 0. The value of $x$ indicates the phase of the global cycle. The phases of all spiking neurons in the network are with respect to this global cycle.

\begin{figure}[tbp] 
   \centering
   \includegraphics[width=15cm]{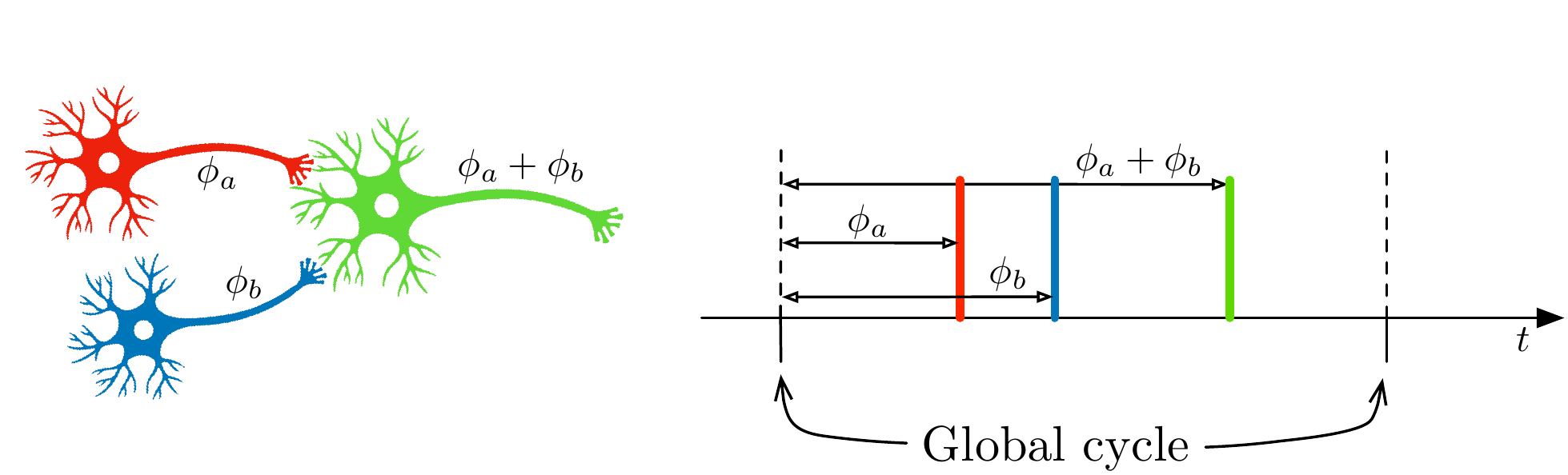} 
   \caption{Phase-sum neuron (in green)}
   \label{fig:phase_sum}
\end{figure}

Let us denote the phases of the incoming spikes as $\phi_a$ and $\phi_b$, each in the range $[0, 2\pi]$. Without loss of generality, assume that $0 \le \phi_a \le \phi_b < 2\pi$. Figure~\ref{fig:phase_sum} shows two neurons projecting to a phase-sum neuron, as well as the timeline showing that the phase-sum neuron fires a spike at phase $\phi_a + \phi_b$.

The general idea behind the phase-sum neuron is that it has an integrator that counts up from the beginning of the cycle until the arrival of the first spike. After the first spike, it holds the integrated value until the arrival of the second spike, at which point it counts down (integrates down) until it reaches 0. 

\begin{figure}[tbp] 
   \centering
   \includegraphics[width=10cm]{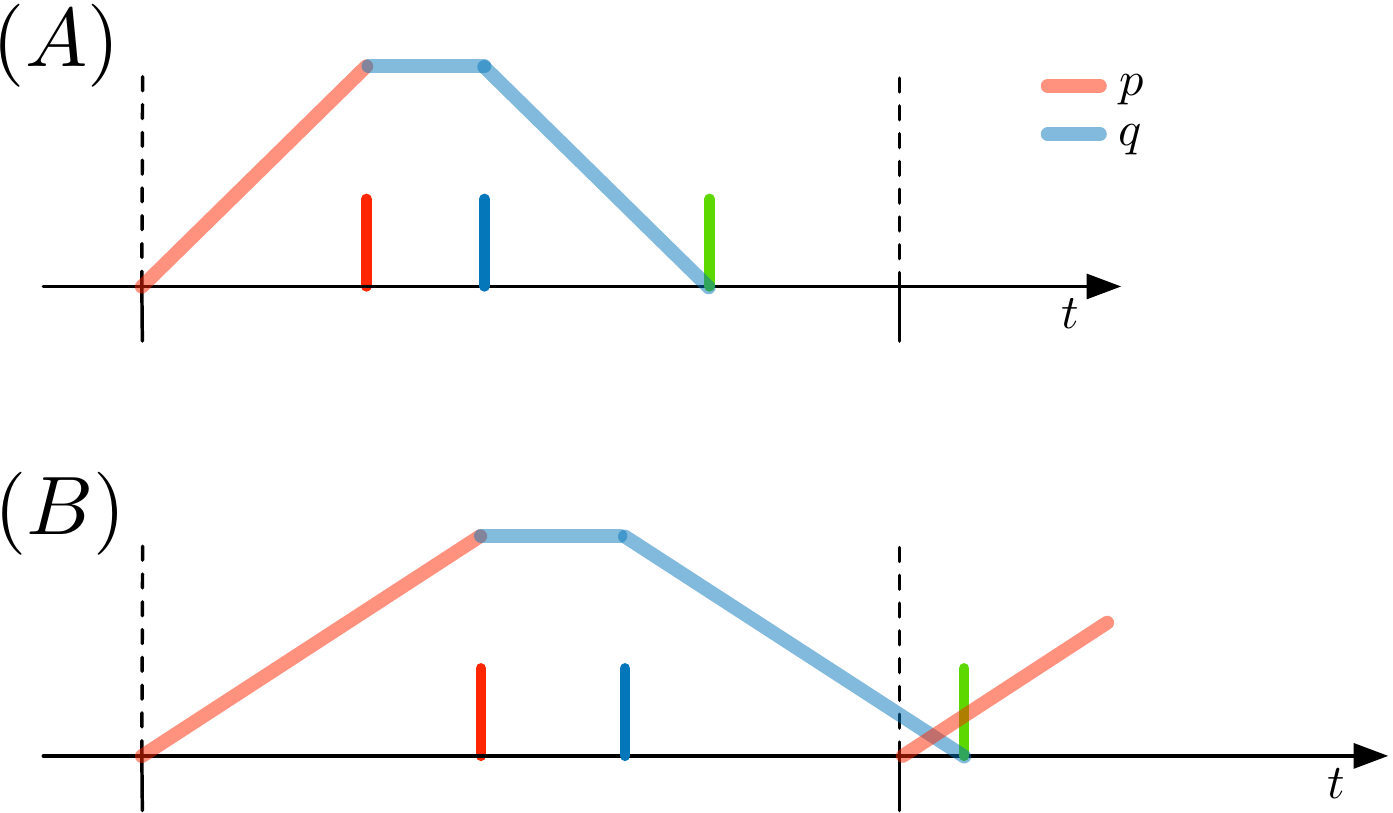} 
   \caption{Phase-sum integrators. At the start of the cycle, $p$ (pink) increases until the first spike arrives. At that point, $q$ (blue) takes on the value of $p$. After the second spike arrives, $q$ decreases until it reaches 0, at which point a spike is generated. In (A), the phase sum is less than the global period. However, in (B), the phase sum is greater than the global period, so $q$ is still counting down at the beginning of the next cycle, when $p$ has to start counting up. The values of $p$ and $q$ are shown only where they are relevant to this cycle.}
   \label{fig:phase_sum_int}
\end{figure}

In actuality, the phase addition is accomplished using a combination of \emph{two} integrators\footnote{It can be done with just one integrator, except when the phase sum is greater than one period, and the computations for adjacent periods overlap.}. Internal variables $p$ and $q$ act as timers. At the beginning of the cycle, $p=0$. As time progresses, $p$ increases according to the differential equation $p' = 1$ until the first spike arrives. At the first spike, we set $q=p$, and $q$ is held constant as time continues, with $q' = 0$. When the second spike arrives, $q$ starts to decrease, following $q' = -1$. Once $q=0$, a spike is generated. This process is illustrated in Fig.~\ref{fig:phase_sum_int}(a). The algorithm for the phase-sum model is articulated in Algorithm~\ref{alg:phase_sum}.

If the sum of the phases is greater than the period of the global cycle, then the computations for adjacent cycles overlap; $p$ has to start counting before $q$ has reached 0. This scenario is shown in Fig.~\ref{fig:phase_sum_int}(b). This overlap is why we use two integrators, instead of one, allowing $p$ to start timing the arrival of the first spike while $q$ is still finishing the calculation from the previous cycle.

\begin{algorithm}
\caption{Event-based computation of phase summation.}
\label{alg:phase_sum}
\begin{algorithmic}
    \State {\bf At start of cycle:} $p \gets 0$, and $p'=1$
    \State {\bf At arrival of first spike:} $q \gets p$, and $q'=0$
    \State {\bf At arrival of second spike:} $q'=-1$
    \If{$q<0$} 
        \State {\bf SPIKE}, and $q \gets 0$, $q'=0$
    \EndIf
\end{algorithmic}
\end{algorithm}

\subsection{Phase-Subtraction Neuron Model}

In the FHRR, unbinding is the same as binding, but by the \emph{conjugate} of the vector; this is accomplished by a Hadamard product of the form $\mye{\phi_a} \e^{-\i \phi_b}$, which is $\mye{(\phi_a - \phi_b)}$. Hence, unbinding is the same as subtracting phases. The phase-subtraction neuron computes the time that elapses from when the spike arrives from $b$ to when the spike arrives from $a$. This is the only neuron model for which the incoming synapses have to be distinguished from each other; this is a natural consequence of the fact that unbinding is not a commutative operation.

Similar to the phase-sum model, the phase-subtraction model has a global-cycle integrator, $x$. It also has two other integrators that work together to generate a spike at phase $\phi_a - \phi_b$.

\begin{figure}[tbp] 
   \centering
   \includegraphics[width=10cm]{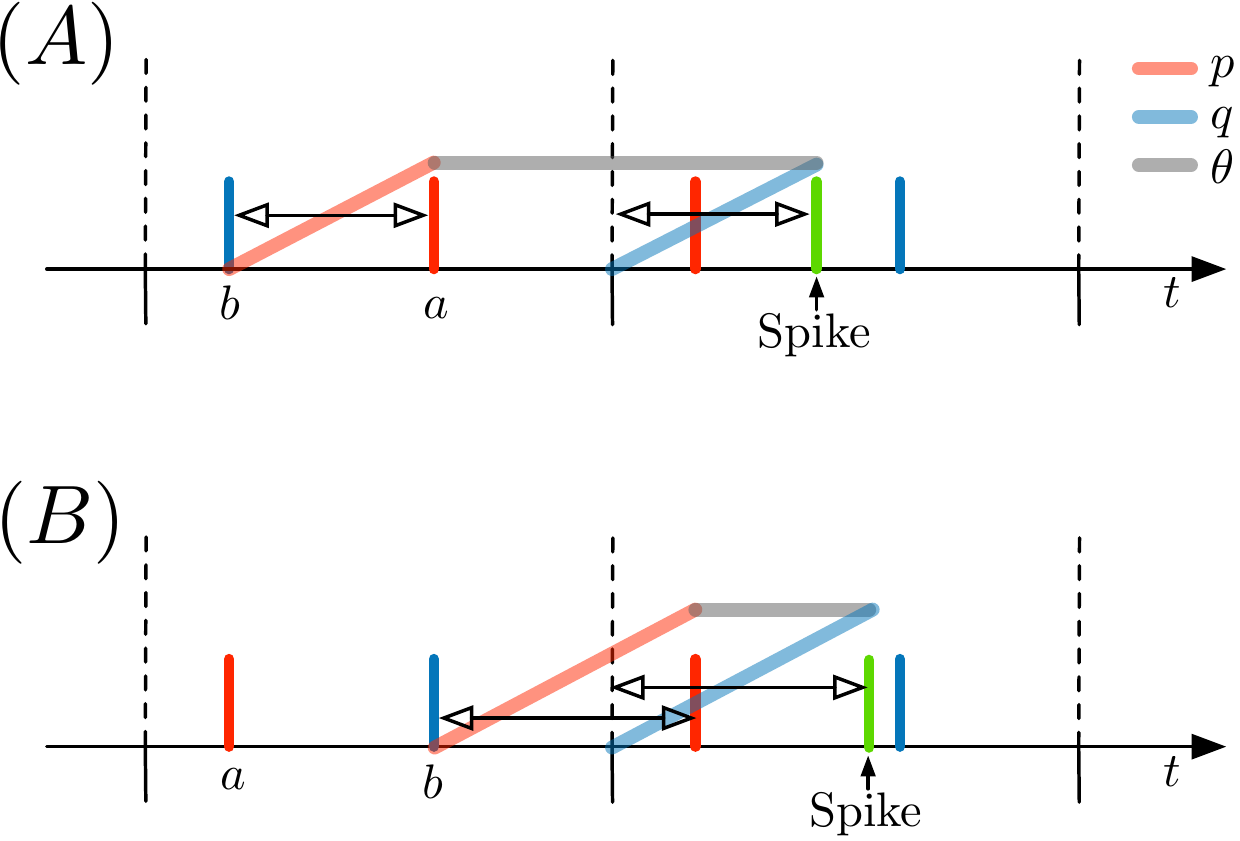} 
   \caption{Phase-subtraction integrators. After the spike from input neuron $b$, the value of $p$ (pink) increases until the spike arrives from $a$. At that point, the threshold value $\thresh$ (grey) is set to $p$. At the beginning of the cycle, $q$ (blue) increases until it reaches the threshold $\thresh$, at which point the neuron generates a spike. In (A), the spike from $b$ arrives earlier (in the cycle) than the spike from $a$. In (B), the opposite is true. In each of (A) and (B), the horizontal arrows show equal spacing. The values of $p$, $q$, and $\thresh$ are shown only where they are relevant to this cycle.}
   \label{fig:phase_sub}
\end{figure}

When the spike arrives from $b$, the variable $p$ is set to 0, and starts integrating according to $p' = 1$. When the spike arrives from $a$, a threshold variable is set, $\thresh = p$, essentially recording the elapsed time between the two spikes. Note that this process might span across two adjacent periods of the global cycle. In parallel, the integrator $q$ is set to 0 at the start of each global cycle, and then increases according to $q' = 1$. When $q$ reaches the threshold, $\thresh$, the neuron spikes. These processes are depicted in Fig.~\ref{fig:phase_sub}.

\subsection{Phase-Multiplication Neuron Model}
\label{sec:phase-mult}

Fractional binding is accomplished by multiplying the phase. We can encode the value $\alpha$ in the complex number $\mye{\phi}$ using $\mye{\phi \alpha}$. If the phase of the input spike was $\phi$, then the phase of the resulting spike is $\phi \alpha$. Note that the input phases are interpreted in the range $[-\pi, \pi]$ instead of $[0, 2\pi ]$. For example, if $0<\alpha<1$, and $\phi = \frac{\pi}{3}$, then the resulting spike will occur earlier in the cycle.

This neuron model has a baseline cycle integrator, $\hat{x}$. Similar to how the phase $\phi$ is centred on 0, the cycle integrator goes from $-\frac12$ to $\frac12$ (instead of 0 to 1), but reaches 0 at the same time as the global cycle. Its evolution is governed by the differential equation $\hat{x}' = 1$. Another integrator, $\hat{p}$, also starts the cycle at $-\frac12$, and increases, governed by $\hat{p}' = 1$.

\begin{figure}[tbp] 
   \centering
   \includegraphics[width=12cm]{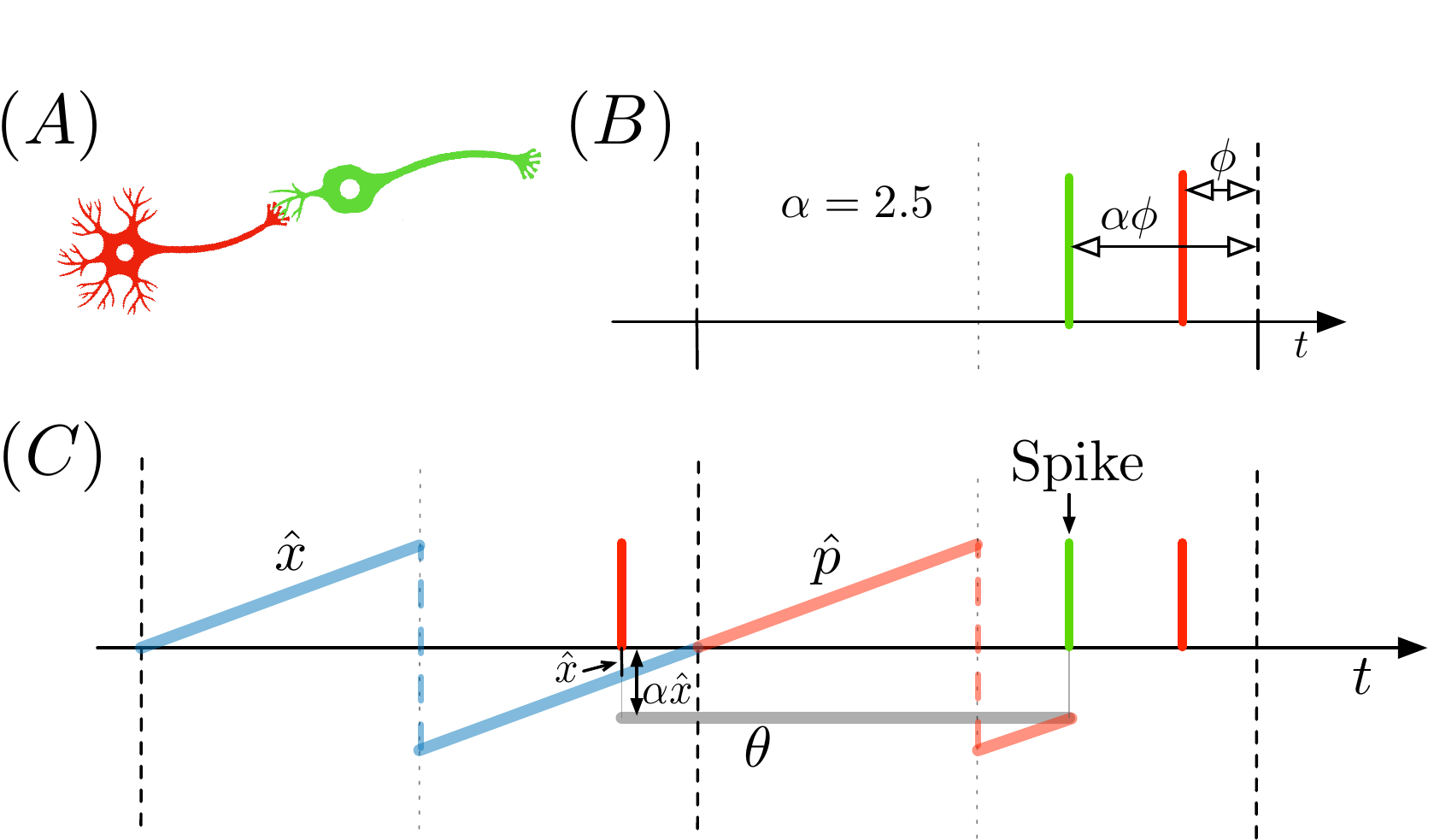} 
   \caption{Phase-multiplication neuron model. (A) A single input neuron projects to the phase-multiplication neuron. (B) The phase of the input spike, $\phi \in (-\pi,\pi]$, is multiplied by $\alpha$ (which is 2.5 in this example), resulting in a spike with phase $\alpha \phi$. (C) When the spike arrives, the threshold $\thresh$ is set to $\alpha \hat{x}$. A spike is generated (in the next cycle) when the integrator $\hat{p}$ reaches $\thresh$. The values of $\hat{x}$, $\hat{p}$, and $\thresh$ are shown only where they are relevant to this cycle.}
   \label{fig:phase_mult}
\end{figure}

When a spike arrives (from the single input neuron), the threshold $\thresh$ is set to $\alpha \hat{x}$. The neuron generates a spike when $\hat{p}>\thresh$. This process is illustrated in Fig.~\ref{fig:phase_mult}.

\subsection{Phase-Averaging Neuron Model}

In the FHRR, bundling is done by adding the vectors. However, adding unitary vectors usually does not result in a unitary vector. To convert the resulting vector to a unitary vector, the modulus information is discarded. However, the phase still contains some information that can be used.

When adding two complex numbers with the same modulus, the phase of the resulting complex number is the average of the input phases. If the input spikes arrive at 0.2\,s and 0.5\,s, then the spike of the bundling neuron should occur at 0.35\,s. Care should be taken to interpret the incoming phases in the range $[-\pi, \pi]$, rather than $[0, 2\pi]$, so that it yields phases closer to 0.

The resulting complex number has to be represented using a spiking phasor neuron, so only the phase of the complex sum is encoded in the neuron.

The time of the average phase is determine using two integrators. Let $p$ be an integrator that starts integrating after the arrival of the first spike, and $q$ be an integrator that starts after the second spike. These integrators reach a value of 1 after one period of the baseline cycle (just as its next spike arrives from the same input neuron). Hence, $(1-p)$ can be thought of as a count-down integrator, essentially tracking the time until the arrival of the next input spike. The midpoint between the 2 spikes is the time when $p=1-q$, or $1-p=q$. But since the furthest apart two spikes can be is half a cycle, then half of that must be less than $\frac14$ of a cycle. Hence, the phase-averaging spike should not be more than $\frac14$ cycle from either of the incoming spikes. Using this logic, the phase-averaging neuron spikes when
$$
\begin{array}{lll}
p\ge1-q & \text{and} & p\le \frac14 \, ,  \text{ or} \vspace{1mm} \\
q\ge 1-p & \text{and} & q\le \frac14 \ .
\end{array}
$$

\subsection{Permutation}

This is simple; we just re-order the vector components (neurons), usually by circularly shifting all the vector components of $v$ by 1 (so that  $\rho(v)_k = v_{k+1}$). This new (shifted) vector, $\rho(v)$, is almost orthogonal to the original, $v$, with a cosine similarity close to zero ($\rho(v) \odot v \approx 0$). Recall that their phases are chosen randomly. Hence, the product of the vector with its shifted counterpart will yield complex numbers with phases that are uniformly distributed. Adding up all those components will -- on average -- be close to zero.

The inverse operation is just as easy. Simply shift the elements back to their original positions.

\subsection{Clean-Up Memory}

The clean-up memory consists of 2 populations, $G$ and $H$, each containing resonate-and-fire (RF) neurons. The goal of the clean-up memory is to have the neurons in $G$ converge to the nearest vector in the vocabulary. Suppose the vectors in our VSA are $N$-dimensional, so that the population $G$ has $N$ RF neurons. That population projects to $H$, which is a population of $M$ RF neurons, where $M$ is the number of patterns we have in our vocabulary. The connection-weight matrix, $W\in\mathbb{C}^{N \times M}$, between $G$ and $H$ contains the vocabulary vectors, one pattern in each column. The input to $H$ can be thought of as $W^\mathrm{T} g$, where $g \in \mathbb{C}^N$ is the unitary complex vector representing the spiking phases of the neurons in $G$. This input is like taking the dot-product of $g$ with each vocabulary vector. If $g$ is close to the vector with index $k$, then we would expect the input to $h_k$ to be larger than the input to the other hidden neurons. In this sense, each neuron in $H$ represents one of the vocabulary vectors. Importantly, this complex matrix-vector product is implemented using scalar weights and spike delays, as described in \citep{Frady2019}.

The neurons in $H$ all inhibit each other; each time a neuron spikes, it sends strong, inhibitory input to all the other neurons in $H$. Because of this mutual inhibition, the population tends towards a winner-takes-all (one-hot) state.

Finally, $H$ projects back to $G$ using connection weights $W$. That is, each time a neuron in $H$ spikes, it sends to $G$ a weighted copy of the pattern it represents (again, using scalar weights and synaptic delays).


\section{Experiments}

We implemented the spiking neuron models in the Brian2 \citep{Stimberg2019}. We then tested our spiking-phasor VSA framework on a number of tasks, two of which we describe here.

\subsection{State Transition Model}

\begin{figure}[tbp] 
   \centering
   \includegraphics[width=5cm]{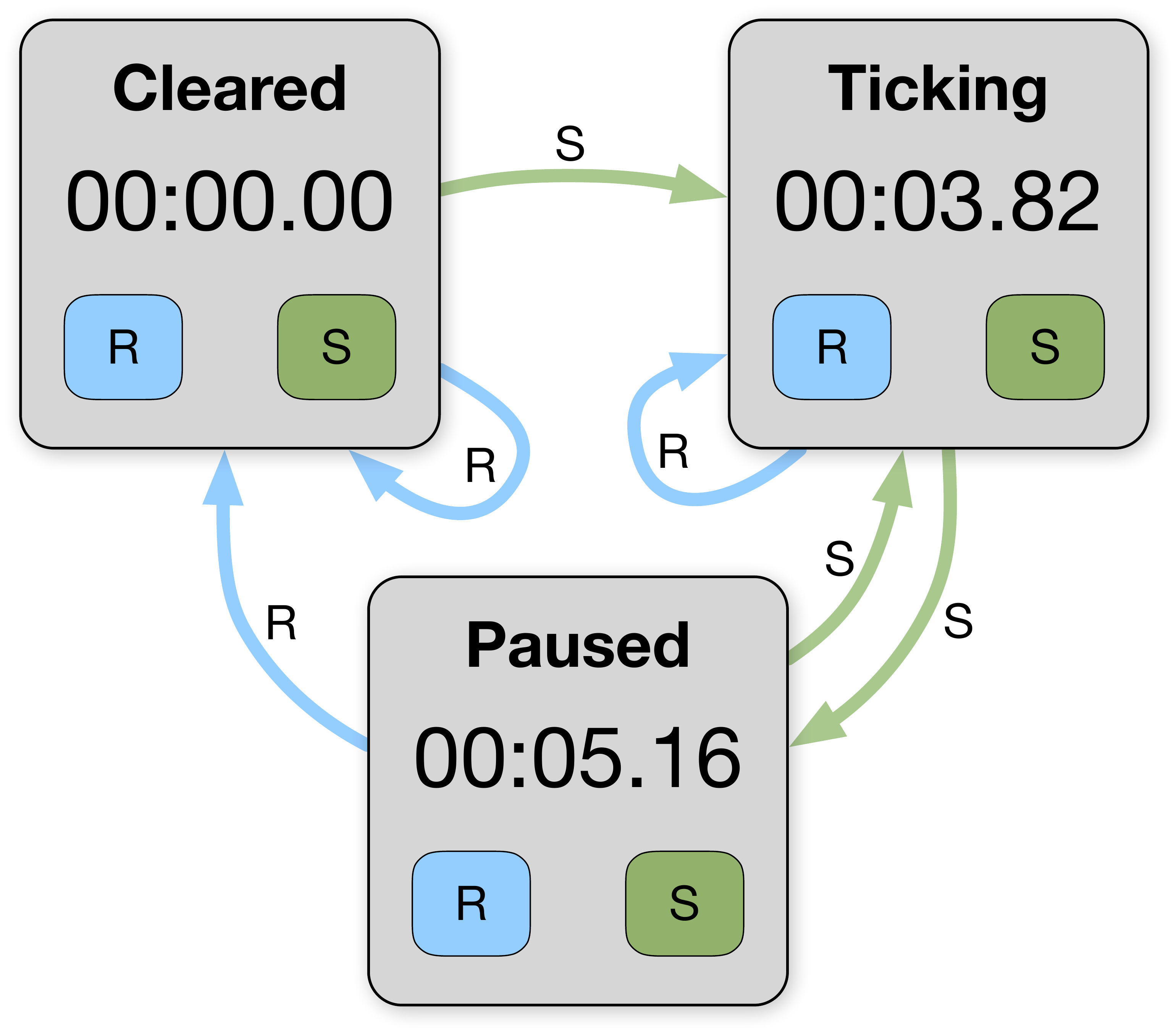} 
   \caption{Stopwatch state transition model. The stopwatch has two buttons: Start/Stop, and Reset/Record.}
   \label{fig:stopwatch}
\end{figure}

Figure \ref{fig:stopwatch} illustrates 3 basic states of a stopwatch, \emph{Cleared}, \emph{Ticking}, and \emph{Paused}. The stopwatch also has 2 buttons, one labelled ``R'', which stands for \emph{Reset/Record}, and one labelled ``S'', which stands for \emph{Start/Stop}. Pressing these buttons can cause the state to transition to a different state. For example, when the stopwatch is in the Cleared state and the S button is pressed (Start), the stopwatch moves to the Ticking state. And when in the Ticking state, pressing the R button keeps it in the Ticking state (to Record a time stamp).

If we represent each of the 3 states with vectors, denoted $C$, $T$, and $P$, and the two actions as vectors, denoted $R$ and $S$, then we can encode a state transition as a vector. For example, a vector $f$ that represent the transition from Cleared to Ticking when the S button is pressed can be constructed using $f = C \otimes S \otimes \rho (T)$. Binding with the vector $\rho (T)$ indicates that the resulting state is Ticking. We can then query $f$ using a vector $q$ of the form $q=\neg C \otimes \neg S$, which asks, ``What state will the stopwatch be in if it is in the Cleared state and we press the S button?'' The resulting vector will be close to $\rho (T)$, a permuted version of $T$. We unpermute it using $\rho^{-1}$ to get the resulting vector $T$.

There are 6 different state transitions, in total. We can bundle all 6 into a single vector using
\begin{align}
f = & \left( C \otimes R \otimes \rho (C) \right) \oplus
 \left( C \otimes S \otimes \rho (T) \right) \oplus
 \left( T \otimes R \otimes \rho (T) \right) \oplus \nonumber \\
 & \left( T \otimes S \otimes \rho (P) \right) \oplus
 \left( P \otimes R \otimes \rho (C) \right) \oplus
 \left( P \otimes S \otimes \rho (T) \right) \label{eq:transitions}
\end{align}
where the $\oplus$ symbol represents bundling, simple vector addition in the case of an FHRR.

We created a 100-dimensional VSA and randomly chose vectors to represent the states and actions, $\{C, T, P, R, S\}$. Each random vector was constructed by randomly choosing 100 phase angles uniformly from the interval $[0, 2\pi)$. The vector $f$ was created by adding together the complex-valued vectors in (\ref{eq:transitions}).

\begin{figure}[tbp] 
   \centering
   \includegraphics[width=10cm]{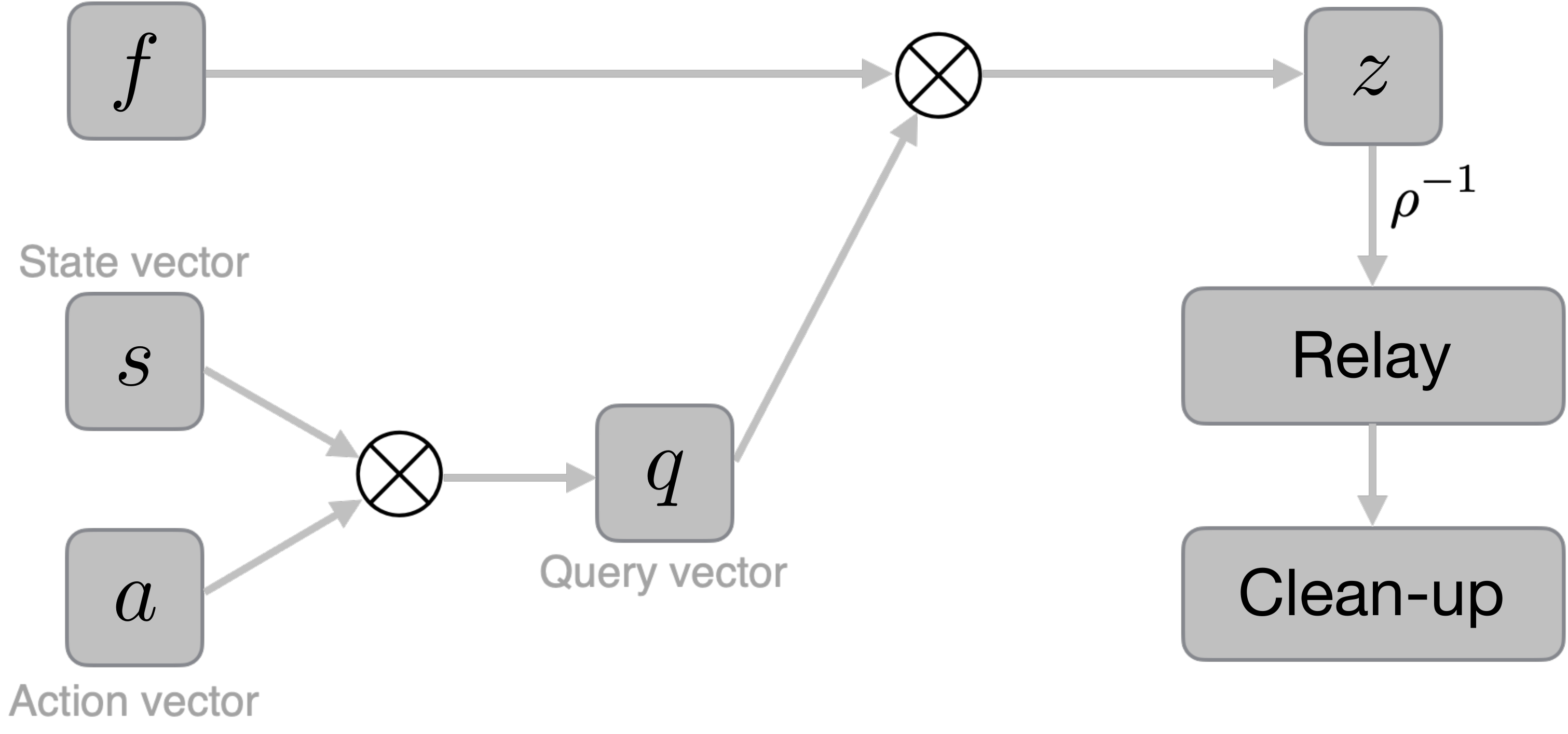} 
   \caption{State transition network. Each gray box represents a population of 100 neurons, except for the Clean-up node, which consists of 105 neurons. All the state transitions are encoded in $f$, and the desired state and action are set in $s$ and $a$, respectively. The population $q$ and $z$ are each 100 phase-sum neurons. The }
   \label{fig:stopwatch_network}
\end{figure}

A population, $F$, was used to store $f$. Each of the 100 neurons in $F$ was firing at the baseline frequency of 10\,Hz. The phases of the elements of $f$ were used to set the spike timing of the 100 neurons in $F$. Figure \ref{fig:stopwatch_network} shows the full network, consisting of a total of 705 spiking neurons. The user chooses the state and action, and the output is read from the clean-up memory.

\begin{figure}[tbp] 
   \centering
   \includegraphics[width=10cm]{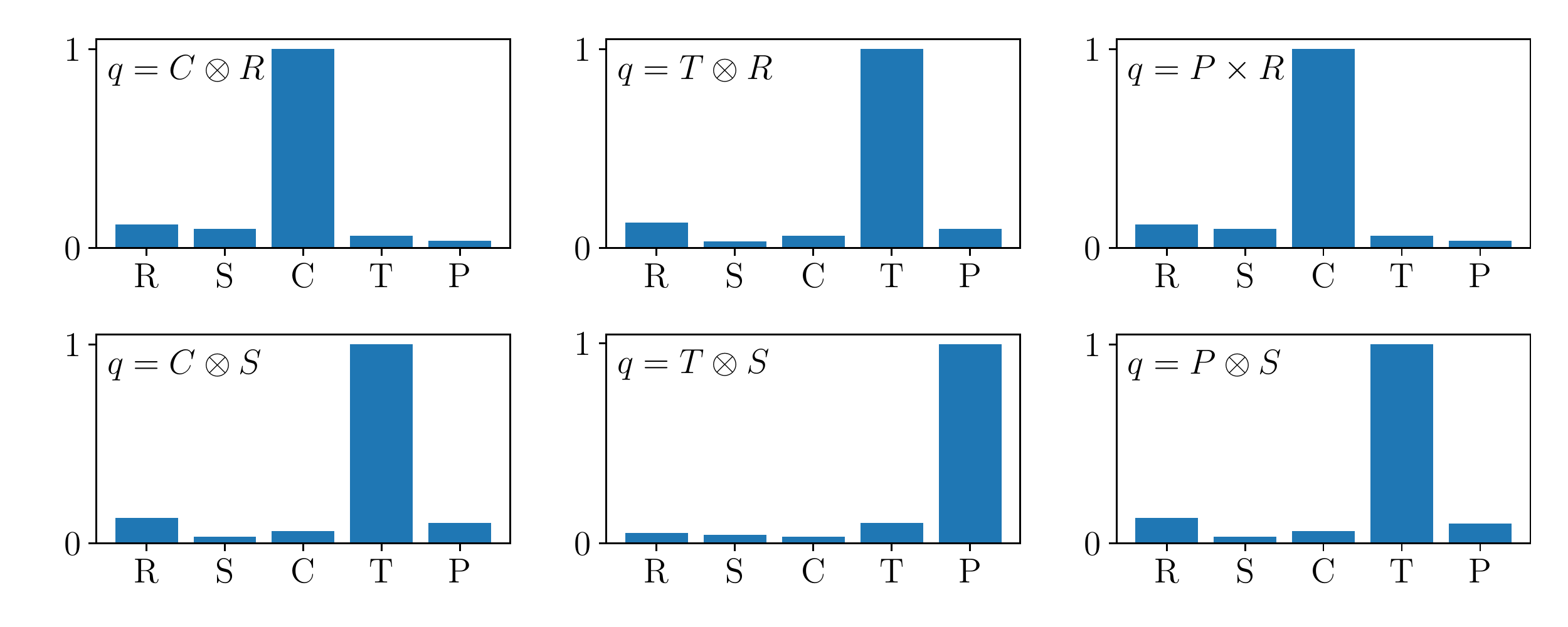} 
   \caption{Results from the state transition network. Each plot shows the query in the top-left, and the bars plot the similarity of the network's output to each of the 5 vectors in the vocabulary.}
   \label{fig:stopwatch_results}
\end{figure}

The results of the stopwatch state transition network are shown in Fig.~\ref{fig:stopwatch_results}. In all cases, the network produced the correct state with a confidence in excess of 99\%.

\subsection{Spatial Memory Model}

The second experiment involves encoding objects and their locations. For this experiment, we will use Spatial Semantic Pointers (SSPs), made possible using fractional exponents in the FHRR. As described in section~\ref{sec:phase-mult}, we can raise our FHRR vectors to fractional exponents, and thereby encode that exponent in the resulting vector. This method has been used to encode spatial position \citep{Frady2018,Lu2019,Dumont2019}.

We randomly generated some 200-dimensional unitary vectors to represent the properties \emph{Square}, \emph{Circle}, \emph{Red}, and \emph{Blue}. These vectors were generated by randomly sampling 200 phases from a uniform distribution over $[0, 2\pi)$. In addition, we used the same method to create another random vector, $X$, to represent a spatial axis. Then we computed a vector that contained multiple objects at different positions,
$$
v = \left( \text{Red} \otimes \text{Square} \otimes X^{1.85} \right) \oplus \left( \text{Blue} \otimes \text{Circle} \otimes X^{-0.65} \right) \ .
$$
The vector $v$ represents a red square at location 1.85, and a blue circle at location -0.65.

We queried $v$ by unbinding it from $X^{1.85}$, essentially asking ``What is at location 1.85?'' The output from the unbinding was fed into the clean-up memory. The state of the neurons in $G$ were compared to the vocabulary vectors, which included: \emph{Square}, \emph{Circle}, \emph{Red}, \emph{Blue}, $X$, $Y$, \emph{Red}$\bind$\emph{Square}, \emph{Blue}$\bind$\emph{Circle}, \emph{Red}$\bind$\emph{Circle}, \emph{Blue}$\bind$\emph{Square}. The closest vocabulary vector was \emph{Red}$\bind$\emph{Square}, with a similarity of 0.999.

\begin{figure}[tbp] 
   \centering
   \includegraphics[width=10cm]{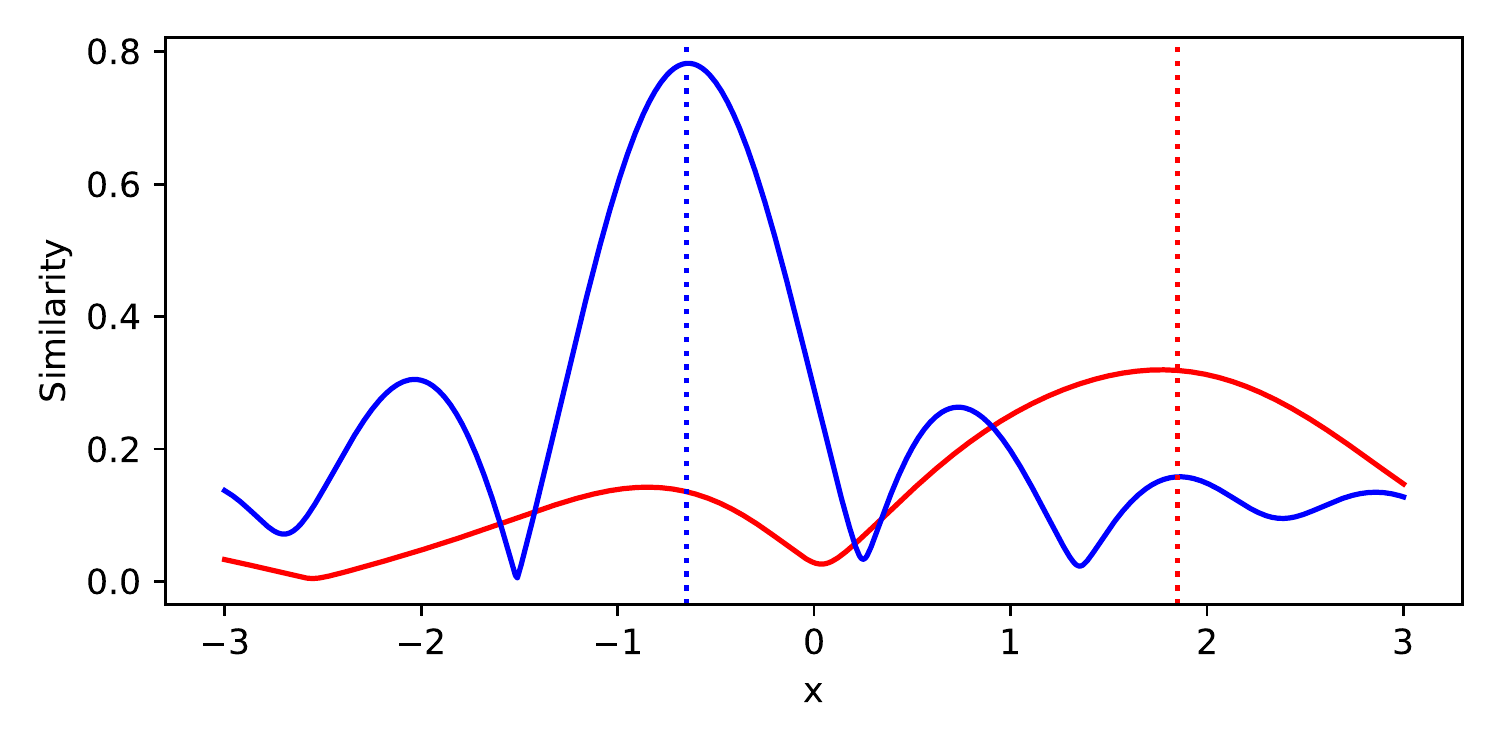} 
   \caption{Similarity of $q_1$ (in red) and $q_2$ (in blue) to $X^x$. Notice that the peak of $q_1 \sim X^x$ occurs close 1.85 (indicated by the dotted red line), and the peak of $q_2 \sim X^2$ occurs at -0.65 (indicated by the dotted blue line).}
   \label{fig:SSP_plots}
\end{figure}

We also queried $v$ by unbinding it from \emph{Red}$\bind$\emph{Square}, and \emph{Blue}$\bind$\emph{Circle}, essentially asking ``Where is the red square?'', and ``Where is the blue circle?''
\begin{align*}
q_1 &= v \oslash \left( \emph{Red}\bind\emph{Square} \right) \\
q_2 &= v \oslash \left( \emph{Blue}\bind\emph{Circle} \right)
\end{align*}
The resulting populations of spiking neurons were converted to complex vectors and their similarity was calculated against a range of SSPs of the form $X^x$, for $-3 \le x \le 3$. Figure~\ref{fig:SSP_plots} plots the similarity over this range of $x$ values. The entire neural network, including all 3 queries, was implemented using 3,406 spiking neurons.

\section{Conclusions}

This paper describes our method for implementing FHRR using spiking-phasor neurons and simple integrator neurons. It extends the ideas proposed in \citep{Frady2019} to operators other than binding. The various vector operations have fairly straight-forward implementations, using integrators to compute the required spike times.

The functionality of these spiking FHRR methods was demonstrated on two tasks, a finite state-transition model, and a spatial memory.

A number of open questions stem from this work. Recall that this spiking-phasor implementation of FHRR does not model the modulus of the complex numbers, just the phase. Thus, all modulus information is discarded. This information loss is particularly relevant after a bundling operation, where the theoretical result would not be unit-modulus, but encoding using spiking-phasor neurons necessitates unit-modulus. It would be interesting to investigate the impact of this loss of information more carefully. For example, what impact does the order in which vectors are added to a bundle have? Can the vectors be chosen in a way that minimizes this loss?

Another direction of further development is a clean-up memory for SSPs. As with any vector in a VSA, sequential binding, bundling, and unbinding deposits noise. This is just as true for an SSP. The clean-up memory proposed in this paper works for a discrete set of vocabulary vectors. In contract, an SSP lies on a 1-dimensional manifold in the vector space, parameterized by the exponent. There is no specific vector to converge to. Instead, a clean-up memory for an SSP would enforce constraints on the phases in the SSP, in a way that pushes the vector back onto the 1-D manifold. This type of coupling has been demonstrated \citep{Ji2014}, though not on spiking-phasor neurons.

Finally, we plan to implement these methods to run them on a neuromorphic platform. Options include porting our code to Lava to run on the Loihi chip \citep{Davies2021}, or on an FPGA platform \citep{Wang2014}.

\bibliographystyle{Frontiers-Harvard} 
\bibliography{refs}


\end{document}